\documentclass[pmlr,twocolumn,10pt]{jmlr}

\mlhtrack{findings}

\newif\iffinal
\finaltrue

\iffinal
  \ifmlhneedspmlr
    \jmlrvolume{XXX}
    \jmlryear{2026}
  \fi
  \ifmlhfindings \jmlrproceedings{}{ML4H 2026 - Findings Track}\fi
  \ifmlhdemo     \jmlrproceedings{}{ML4H 2026 - Demo Track}\fi
  \jmlrworkshop{Machine Learning for Health (ML4H) 2026}
\else
  \jmlrproceedings{}{Submitted to ML4H 2026: \mlhtrackname}
  \jmlrworkshop{Machine Learning for Health (ML4H) 2026}
\fi

\usepackage{booktabs}
\usepackage[switch]{lineno}
\usepackage{placeins}
\usepackage{url}
\usepackage{makecell}

\title[Auditing Attribution-Driven Curation]{Leakage-Safe Auditing of Attribution-Driven Data Curation in Medical Imaging}
\author{\Name{Malay Phadke}\Email{malay.phadke@spit.ac.in}\\
\addr Sardar Patel Institute of Technology, Mumbai, India
\AND
\Name{Akshay Gokhale}\Email{akshay.gokhale22@spit.ac.in}\\
\addr Sardar Patel Institute of Technology, Mumbai, India
}

\begin{document}
\maketitle

\begin{abstract}
When can an attribution-derived anomaly justify changing a medical-imaging
training set? We formulate attribution-driven curation as three separately
testable claims---anomaly detection, harmfulness estimation, and intervention
utility---and test the first and third with a patient-disjoint,
identifier-preserving audit of a 3D convolutional classifier for
computed-tomography-angiography aneurysm candidates. The audit identifies
232 exactly zero-valued arrays among 1,201 positive patches, including 170
training positives, and invalidates a broader historical 321-patch
attention-failure subgroup whose identifier-free records had been joined to
files by position. On the held-out test split, GradCAM++ spatial
concentration is central for every candidate class, reversing the historical
classwise pattern; GradCAM++ and Integrated Gradients show negative
rank-order agreement for positives, preventing a method-independent
anatomical interpretation. Across five paired training seeds, removing the
170 corrupted training positives yields AUC $0.9617\pm0.0044$ versus
$0.9652\pm0.0039$ for full-data training (strict-minus-full AUC $-0.0035$;
paired-seed bootstrap 95\% CI $-0.0082$ to $0.0007$); removing the 34
raw-zero test positives widens the gap rather than closing it, contradicting
a blank-memorization explanation. Budget-matched random-removal controls
confirm the result is not an artifact of reduced training size. Genuine input
corruption need not imply beneficial deletion. The framework makes leakage
control, sample identity, and intervention testing explicit components of
interpretability-based curation in medical-imaging pipelines.
\end{abstract}


\ifmlhneedsstatements
\paragraph*{Data and Code Availability}
This study uses previously prepared CTA-derived patch artifacts under
applicable data-use restrictions; source clinical volumes cannot be
redistributed. Code for the audit pipeline, curation experiments, and
attribution analyses is available at \url{https://github.com/MalayPhadke/ml4h_aneursym_audit}.
The appendix reports the split construction, executable settings, artifact
hashes, and derived non-identifying results needed to audit the analyses.

\paragraph*{Institutional Review Board (IRB)}
This study performs secondary analysis of previously prepared, de-identified
clinical-imaging-derived patch artifacts with no new data collection, patient
contact, or re-identification. IRB review was not required under the
applicable institutional policy for analysis of existing de-identified data.
\fi

\section{Introduction}
\label{sec:introduction}

Attribution maps offer an attractive route from model interpretation to data
curation: score each image, inspect suspicious cases, and remove or repair
those judged anomalous. That workflow nevertheless combines distinct
inferences. A score can detect an unusual sample without showing that the
sample harms learning, and harmfulness does not guarantee that deletion is the
best intervention. Conflating these claims is consequential in clinical AI,
where a plausible heatmap may motivate changes to a limited or clinically
important training subset.

Medical-imaging pipelines create particular audit risks. Multiple patches
can derive from one patient, so patch-wise splitting can contaminate
evaluation. Candidate generation, preprocessing, inference, attribution,
and clinical metadata are often executed separately and joined later;
losing sample identity at any interface can attach an explanation to the
wrong image \citep{geirhos2020,adebayo2018,saporta2022}.

We audit a specific instance of this risk. Historical internal analyses of the
same pipeline reported peripheral GradCAM++ concentration in positive patches
and identified a 321-patch attention-failure subgroup. Those analyses were not
published; the present study is the first leakage-safe re-evaluation.

We therefore ask: \emph{when does an attribution-derived anomaly justify a
training-data intervention?} We decompose the question into three empirical
claims: \textbf{anomaly detection} (does the signal identify abnormal
samples?), \textbf{harmfulness estimation} (do those samples adversely affect
learning or generalization?), and \textbf{intervention utility} (does the
proposed deletion improve fixed held-out performance?). Success on one claim
does not establish the next. This paper tests anomaly detection and
intervention utility; harmfulness estimation remains an open link.

Our case study is a two-stage intracranial aneurysm (IA) candidate pipeline
for computed tomography angiography (CTA), in which vesselness-derived
candidate locations are represented by $64^3$ patches and classified by a 3D
convolutional neural network (CNN). This study is an audit of a candidate
classifier, not an estimate of end-to-end diagnostic performance. The audit
tests whether historical attribution-based observations reproduce under
corrected methodology; it does not evaluate whether attribution maps are
faithful to model reasoning. The novelty lies in the audit construction and
empirical demonstration, not in the component methods individually.

\section{Related Work}
\label{sec:related}

\paragraph{IA candidate classification.}
Candidate-generation followed by patch classification is established for
intracranial aneurysm detection
\citep{frangi1998,lauric2010,nakao2018}. The two-stage design
introduces a hard interface between candidate extraction and learned
classification: every downstream evaluation inherits the identity and
quality of the extracted patch. Patient-disjoint splitting is necessary
whenever multiple patches derive from one patient, yet it is not
universally enforced in patch-level studies.

\paragraph{Dataset auditing and curation.}
Confident Learning can flag likely label errors at scale
\citep{northcutt2021}, and data pruning can remove difficult or redundant
examples to improve training efficiency \citep{paul2021}. However, an
accurate anomaly detector is not automatically an effective deletion
policy: removing a detected anomaly improves performance only if the
anomaly genuinely harms learning and the removal does not introduce a
worse bias.

\paragraph{Attribution methods and reliability.}
Grad-CAM \citep{selvaraju2017}, GradCAM++ \citep{chattopadhay2018}, and
Integrated Gradients \citep{sundararajan2017} provide gradient-based
views of model predictions. Saliency maps can fail basic sensitivity and
implementation checks \citep{adebayo2018}, and clinical deployment
studies report disagreement between methods and between methods and
expert annotation \citep{saporta2022}. Cross-method disagreement weakens
any method-independent anatomical interpretation drawn from a single
attribution technique.

\section{Audit Framework and Experimental Design}
\label{sec:methods}

\subsection{Three separately testable claims}

The audit follows a fixed sequence: an anomaly score is linked to raw
input by immutable patch identifier and checked against a direct quality
criterion; potential harmfulness is assessed without consulting test
labels; intervention utility is evaluated by retraining against both an
untouched full-data condition and a budget-matched random-removal condition.Patient separation and identifier integrity are enforced before model training, not verified after results are available.

\subsection{Clinical task and dataset}

The frozen dataset derives from the RSNA 2025 Intracranial Aneurysm
Detection Challenge~\citep{rsna2025}, a multi-institutional CTA collection
spanning four scanner manufacturers. It contains 7,742 CTA-derived patches
from 1,793 patient/scan groups: 1,201 positives and 6,541 negatives across three negative
subclasses---Frangi negatives (candidate locations triggered by the Frangi
vesselness filter \citep{frangi1998} that are not true aneurysms), hard
negatives, and easy negatives. A seeded, stratified group split assigns patches to
training (1,256 patients), validation (270), and test (267) with zero
pairwise patient overlap. Complete counts appear in
Appendix~\ref{app:dataset}.

\subsection{Evaluation, model, and preprocessing}

All reported models use a 3D MONAI ResNet-10 ($64^3$ input), trained from
scratch for 20 epochs with Adam ($\text{LR}=10^{-4}$), unweighted cross
entropy, and no augmentation. The validation checkpoint minimizes FPR@90;
the fixed test split is evaluated only after selection. A balanced training
subset contains 3,460 patches. Full specifications appear in
Appendix~\ref{app:model}. Preprocessing clips, smooths, and min--max
normalizes each patch; an identically zero raw array remains zero after
this transform, validating the raw-zero criterion.

\subsection{Attribution and spatial summaries}

GradCAM++ targets the positive-class logit at the last convolutional layer;
IG uses a zero baseline with 100 trapezoidal intervals. Both are computed for
every frozen patch (Appendix~\ref{app:attribution}). We summarize spatial
concentration with the radial attribution ratio (RAR), the ratio of mean
outer-sphere to mean inner-sphere normalized attribution
(Appendix~\ref{app:attribution}). RAR below one indicates central
concentration.

\subsection{Identifier-preserving audit}

Every current attribution row retains its canonical relative file path as
\texttt{patch\_id}, together with patient ID, split, class, and raw-array
path. Joins are keyed by patch ID with one-to-one coverage. Direct
inspection of the serialized arrays defines a strict corruption flag as an
identically zero raw positive patch.

The historical attribution output omitted patch IDs; a downstream analysis
paired rows with sorted files by position, so row-level identity could not
be verified (Figure~\ref{fig:join-audit} in the appendix). The historical
evaluation also lacked a patient-disjoint held-out sample: its evaluation
data came from a source pool used for training, so the result is not
leakage-safe evidence of generalization.

\subsection{Curation interventions and statistics}

The intervention study compares: (i)~the full baseline with 868 training
positives; (ii)~strict removal of all 170 raw-zero training positives; and
(iii)~matched random removal of a fixed set of 170 randomly chosen training
positives (selection seed 42). All three conditions are trained with five
paired seeds (42--46) to separate the deletion effect from training-seed
variation. A complementary selection-variance control trains five distinct
random-removal subsets under a single training seed
(Appendix~\ref{app:curation}, Table~\ref{tab:app-random170-seeds}). All
other design choices are held constant.

Primary metrics are test AUC and FPR@90. The strict-minus-full AUC
difference uses a paired-seed bootstrap (five seed pairs, 267 test patients
per seed, 20,000 resamples). Statistical methods including enrichment tests
and method-agreement analyses are detailed in Appendix~\ref{app:evaluation}.

\section{Results}
\label{sec:results}

\subsection{Patient-disjoint evaluation reverses the historical classwise pattern}

We recomputed GradCAM++ RAR for the fixed test split using the train-only
checkpoint. All 168 positive test patches have finite GradCAM++ RAR; the
Spearman $\rho$ column is computed over the 134 finite paired GradCAM++/IG
values. The 34 undefined pairs are exactly the raw-zero positives, for
which IG with a zero baseline is identically zero by construction. All four
class medians are below one (Table~\ref{tab:test-attribution}), with
positives more central than Frangi and hard negatives, reversing the
historical classwise pattern. Because the historical evaluation pool was
also used for training, the earlier pattern lacks leakage-safe support.

\begin{table}[t]
  \caption{Held-out test attribution. GradCAM++ RAR is median (patient-cluster
  bootstrap 95\% CI). $\rho$ is test-split Spearman agreement between
  GradCAM++ and IG bounded-density summaries over finite pairs.}
  \label{tab:test-attribution}
  \centering
  \small
  \begin{tabular}{lrrr}
    \toprule
    Class & $n$ & \makecell{GradCAM++ RAR\\(95\% CI)} & $\rho$ \\
    \midrule
    Positive & 168 & 0.669 (0.575--0.722) & $-0.200$ \\
    Frangi & 532 & 0.739 (0.719--0.768) & 0.522 \\
    Hard & 276 & 0.810 (0.798--0.830) & 0.374 \\
    Easy & 160 & 0.703 (0.673--0.736) & 0.382 \\
    \bottomrule
  \end{tabular}
\end{table}
\subsection{Identifier-aware reconstruction undermines the historical subgroup}

Direct audit identifies 232/1,201 positive arrays as exactly zero-valued.
Positional reconstruction associates the other 89 members of the historical
321-patch group with nonzero files, but the identifier-free historical table
prevents a definitive record-level account. The broad 321-record definition
cannot support a trustworthy model-level subgroup; the defensible anomaly is
the 232-patch raw-zero set.

\subsection{Raw-zero patches are enriched among small-lesion positives}

Of the 232 raw-zero positives, 204 (87.9\%) have estimated dome size (the
maximum of the annotated in-plane diameter and z-extent in mm, derived from
NIfTI voxel spacing and DICOM slice positions; for single-slice annotations
(84\%) it reduces to one slice thickness, so it is a coarse,
spacing-limited estimate) below
3\,mm, compared with 76.2\% of nonzero positives. A patient-cluster
bootstrap gives OR 2.29 (95\% CI 1.44--4.13; only 0.015\% of resampled
ORs $\leq 1$), confirming the association at the patient level
(Appendix~\ref{app:evaluation}). This is an association with inherited size
estimates, not evidence that lesion size caused extraction failure.

\subsection{Targeted deletion does not outperform full-data training}

To separate the deletion effect from training-seed variation, all three
conditions were trained with five paired seeds (42--46). Full training
averages AUC $0.9652\pm0.0039$ and strict removal averages AUC
$0.9617\pm0.0044$ (Table~\ref{tab:curation}). The paired strict-minus-full
AUC difference is $-0.0035$ (paired-seed bootstrap 95\% CI $-0.0082$ to
$0.0007$); strict removal has higher AUC in one of five paired seeds. The
matched random-removal condition (fixed selection, five training seeds)
averages AUC $0.9565\pm0.0056$ and FPR@90 $0.0878\pm0.0118$. The
confidence interval includes zero. Targeted deletion shows no reproducible
advantage over full-data training.

As a conservative robustness check, we removed the 34 raw-zero positive
test patches and re-evaluated the seed-42 checkpoints without retraining.
Full-data AUC drops from 0.9671 to 0.9589 (clean FPR@90 0.0692);
strict-removal AUC drops from 0.9589 to 0.9486 (clean FPR@90 0.1167). The
strict model's FPR@90 nearly doubles on the clean test set, while the full
model's remains stable. The strict-minus-full AUC gap widens to $-0.0103$
(bootstrap 95\% CI $-0.0239$ to $0.0006$). This result contradicts the
blank-memorization hypothesis---that full-data performance is inflated by
learning ``blank equals positive.'' Targeted deletion still shows no
advantage under conservative evaluation.

\FloatBarrier
\begin{table}[t]
  \caption{Fixed held-out test performance. All conditions report
  mean$\pm$SD over 5 paired training seeds (selection held constant). Higher
  AUC and lower FPR@90 are better.}
  \label{tab:curation}
  \centering
  \small
  \begin{tabular}{lcc}
    \toprule
    Training condition & AUC $\uparrow$ & FPR@90 $\downarrow$ \\
    \midrule
    Full data (5 seeds) & $\mathbf{0.9652\pm0.0039}$ & $\mathbf{0.0729\pm0.0120}$ \\
    Strict zero removal (5 seeds) & $0.9617\pm0.0044$ & $0.0764\pm0.0088$ \\
    Matched random removal (5 seeds) & $0.9565\pm0.0056$ & $0.0878\pm0.0118$ \\
    \bottomrule
  \end{tabular}
\end{table}

\subsection{Attribution methods lack positive-class convergence}

GradCAM++ and IG bounded-density summaries are negatively associated for
test-set positives ($\rho=-0.200$, $n=134$, $p=0.0206$) but positively for
negatives ($\rho=0.469$, $n=967$). These results prevent treating the two
methods as convergent evidence for the positive-class explanation.

\section{Discussion}
\label{sec:discussion}

The strict rule has high construct validity: every flagged positive is an
identically zero array. Yet the intervention does not improve the full-data
point estimate across five paired training seeds, and the gap widens when
raw-zero test positives are excluded. The reusable boundary condition is
that \emph{anomaly is evidence for inspection, not sufficient evidence for
deletion}. Notably, the zero-array anomaly was found by direct raw-input
inspection, not by the attribution workflow that was intended to detect such
issues; the heatmaps did not flag these patches.

Interpretability validity also depends on identity provenance: immutable
sample identifiers are necessary because an attribution row without a
verifiable relationship to its image cannot sustain sample-level biological
claims. The corrected pipeline treats split membership and join coverage as
auditable artifacts, not implicit properties of file order.

Notably, the defensible anomaly---232 zero-valued arrays---was identified
by direct raw-array inspection, not by the attribution scores themselves.
Attribution motivated the original investigation but did not produce the
quality signal; direct input QC did. This illustrates that attribution can
prioritize where to look without being the mechanism that confirms what is
wrong.
This study does not independently isolate harmfulness: the three-claim
framework is proposed, but only anomaly detection and intervention utility
are empirically evaluated. Whether zero-valued patches actively degrade
learned representations---rather than merely contributing uninformative
gradients---would require influence-function or leave-one-out analyses
beyond the scope of this retrospective audit. Because only positive patches are removed, both strict and random
conditions reduce the effective positive prior under unweighted
cross-entropy. This confound is shared between strict and random arms
(and cancels in their comparison) but is not controlled in the
strict-versus-baseline contrast.

More broadly, candidate-based medical imaging exposes provenance levels from
patient through transformed model input. A credible audit should preserve
these links. The three-claim decomposition clarifies action: direct raw QC
may establish anomaly detection, influence experiments are needed for
harmfulness, and controlled retraining is needed for intervention utility.
Attribution can prioritize inspection but should not bypass those tests.

\section{Limitations}
\label{sec:limitations}

This retrospective audit covers one inherited dataset, one disease, one
preprocessing path, and one model family (MONAI ResNet-10); architecture
dependence remains untested. Source CTA volumes are unavailable, so the
upstream cause of zero patches cannot be reconstructed. The RAR reversal
(Section~\ref{sec:results}) cannot be isolated to the corrected
patient-disjoint split versus the retrained model, because the historical
model checkpoint is not recoverable. Five paired training seeds quantify
seed variation for the deletion comparison; the test-patient bootstrap
captures test-sample uncertainty. The harmfulness link in the three-claim
framework is not directly tested. Training used no augmentation, limiting
generalization claims. Attribution faithfulness was not established with
causal perturbation tests. Lesion-size values are inherited rather than new
blinded measurements. We use patient-cluster analyses where IDs support
them; remaining analyses are descriptive. This patch-classification audit
does not measure end-to-end candidate recall.

\section{Conclusion}
\label{sec:conclusion}

A patient-disjoint, identifier-preserving audit rejected two historical
mechanistic interpretations, isolated a real zero-input anomaly, and found no
held-out benefit from its targeted deletion. The result is that each link
from explanation to action requires separate evidence: attribution-driven
curation should separately establish anomaly, harmfulness, and intervention
utility.

\section*{Ethics statement}

This work audits an AI pipeline for a clinically consequential task and does
not support autonomous diagnosis or deployment. Reported associations are not
clinical risk estimates. Patient imaging must remain subject to applicable
approvals, privacy safeguards, and data-use conditions.

\section*{Use of AI assistants}

Large language models were used for writing assistance during manuscript
preparation and revision. All scientific claims, experimental design,
implementation, analysis, and interpretation are the sole work and
responsibility of the authors.

\bibliography{ref}

\appendix
\renewcommand{\thetable}{A\arabic{table}}
\renewcommand{\thefigure}{A\arabic{figure}}
\setcounter{table}{0}
\setcounter{figure}{0}

\section{Dataset, split, and patch integrity}
\label{app:dataset}

All counts below are recomputed from the frozen manifest rather than
transcribed from the historical scripts. The grouping key is the patient/scan
UID parsed from each patch filename.

\begin{table*}[t]
  \caption{Frozen v1 dataset and patient-grouped split. ``Pos:neg'' reports
  positive patches divided by all negative patches. Patient counts are
  disjoint across rows.}
  \label{tab:app-dataset}
  \centering
  \small
  \begin{tabular}{lrrrrrrrr}
    \toprule
    Split & Patients & Positive & Frangi & Hard & Easy & Total & Pos:neg & Patches/patient median (range) \\
    \midrule
    Train & 1,256 & 868 & 2,506 & 1,261 & 856 & 5,491 & 1:5.33 & 4 (1--27) \\
    Validation & 270 & 165 & 538 & 250 & 162 & 1,115 & 1:5.76 & 2.5 (1--21) \\
    Test & 267 & 168 & 532 & 276 & 160 & 1,136 & 1:5.76 & 4 (2--14) \\
    \midrule
    Total & 1,793 & 1,201 & 3,576 & 1,787 & 1,178 & 7,742 & 1:5.45 & 4 (1--27) \\
    \bottomrule
  \end{tabular}
\end{table*}

The mean number of patches per patient is 4.32 overall (train 4.37,
validation 4.13, test 4.25). The split uses
\texttt{StratifiedGroupKFold} with 20 shuffled folds and seed 42: folds 0--13
form train, 14--16 validation, and 17--19 test. Stratification targets the four
patch classes; grouping is by patient ID. All three pairwise patient-overlap
counts are zero.

\begin{table}[t]
  \caption{Strict raw-zero positive-patch demographics. Strict Mode 2 means
  exact radial slope zero and an exact-zero IG map; Phase-0 raw audit confirmed
  that every listed NPZ \texttt{resized} array is identically zero. ``Small''
  denotes aneurysms $<3$\,mm.}
  \label{tab:app-zero}
  \centering
  \small
  \begin{tabular}{lrrr}
    \toprule
    Split & Patches & Unique patients & Small \\
    \midrule
    Train & 170 & 153 & 146 (85.9\%) \\
    Validation & 28 & 26 & 25 (89.3\%) \\
    Test & 34 & 32 & 33 (97.1\%) \\
    \midrule
    Total & 232 & 211 & 204 (87.9\%) \\
    \bottomrule
  \end{tabular}
\end{table}

Among the 232 strict positives, estimated dome-size categories are 204 small,
13 medium, and 15 large. The median estimated dome size is 1.0\,mm (IQR
0.625--1.0; range 0.6--45.0). Across all four classes, the exact-zero-slope
audit contains 238 patches (232 positive and 6 Frangi), all raw-zero.

\section{Preprocessing and extraction}
\label{app:preprocessing}

Each stored NPZ exposes a $64^3$ array under the key \texttt{resized}. The
reproducible model-input transform first clips and smooths each patch:
\[
z = G_{\sigma=0.5}\!\left(\operatorname{clip}(x,0,800)\right),
\]
followed by min--max normalization:
\[
\tilde{x}
= \frac{z-\min(z)}
       {\max(z)-\min(z)+10^{-6}}.
\]
Here, $G$ is a three-dimensional Gaussian filter and all operations are
performed per patch. This transform maps an all-zero raw array to an all-zero
model input; it cannot create the historical claim of a non-degenerate HU
distribution.

The manifest preserves integer voxel anchors $(x,y,z)$ parsed from filenames
and an immutable relative patch ID. Source CTA volumes and the original v1/v2
extraction scripts are unavailable. Consequently, the Frangi scale range,
vesselness threshold, candidate non-maximum-suppression rule, original voxel
spacing, interpolation used to create \texttt{resized}, physical patch extent,
and boundary-padding rule cannot be verified and are not reconstructed here.
Likewise, the commonly stated centering asymmetry (annotation-centered
positives versus trigger-centered Frangi negatives) is consistent with the
inherited naming and study notes but cannot be independently re-executed from
the released artifacts. This evidence boundary is why paired source-volume
recentering and CAS experiments are reported unavailable.

\section{Model architecture and optimization}
\label{app:model}

\begin{table}[t]
  \caption{Classifier architecture verified from the executable model
  constructor.}
  \label{tab:app-arch}
  \centering
  \small
  \begin{tabular}{p{0.34\linewidth}p{0.56\linewidth}}
    \toprule
    Component & Specification \\
    \midrule
    Backbone & MONAI ResNet-10, 3D, trained from scratch \\
    Input & $1\times64\times64\times64$ \\
    Stem & $7^3$ Conv, 64 channels, stride 1; BN; ReLU; $3^3$ max-pool, stride 2 \\
    Residual stages & One block each at 64, 128, 256, 512 channels; stage strides 1, 2, 2, 2 \\
    Block & Two $3^3$ convolutions with BN and ReLU; $1^3$ projection on downsampling stages \\
    Head & Adaptive global average pooling; linear $512\rightarrow2$ \\
    Parameters & 14,357,442 trainable parameters \\
    Dropout/pretraining & None / none \\
    \bottomrule
  \end{tabular}
\end{table}

\begin{table}[t]
  \caption{Training protocol shared by the full, strict-removal, and
  matched-random conditions.}
  \label{tab:app-train}
  \centering
  \small
  \begin{tabular}{p{0.30\linewidth}p{0.60\linewidth}}
    \toprule
    Setting & Value \\
    \midrule
    Optimizer & Adam \\
    Learning rate / schedule & $10^{-4}$ / constant \\
    Weight decay & $10^{-5}$ \\
    Loss & Unweighted two-class cross entropy \\
    Epochs & 20 (all epochs executed) \\
    Train / evaluation batch & 16 / 32 \\
    Precision & CUDA automatic mixed precision \\
    Seed & 42 \\
    Augmentation & None in the executable loader \\
    Checkpoint selection & Minimum validation FPR\@90; AUC tie-breaker \\
    Test use & Once after validation-only selection \\
    Hardware & NVIDIA RTX PRO 6000 Blackwell Max-Q \\
    \bottomrule
  \end{tabular}
\end{table}

The common baseline training selection uses all 868 train positives, up to 868
seeded examples from each negative subclass, and all 856 available easy
negatives: 3,460 patches total. This is a balanced training subset of the
5,491-patch frozen train pool, not the entire pool.

\begin{table*}[t]
  \caption{Convergence and selection diagnostics for the three
  positive-removal conditions. Every run completed 20 epochs; ``selected
  loss'' is training loss at the validation-selected epoch.}
  \label{tab:app-convergence}
  \centering
  \small
  \begin{tabular}{lrrrrrr}
    \toprule
    Condition & Train $n$ & Positive $n$ & Selected epoch & Selected loss & Validation FPR@90 & Validation AUC \\
    \midrule
    Full-positive & 3,460 & 868 & 8 & 0.1642 & 0.0432 & 0.9709 \\
    Remove strict 170 & 3,290 & 698 & 12 & 0.1342 & 0.0411 & 0.9726 \\
    Remove random 170 & 3,290 & 698 & 11 & 0.1244 & 0.0368 & 0.9655 \\
    \bottomrule
  \end{tabular}
\end{table*}

\begin{figure*}[t]
  \centering
  \includegraphics[width=0.86\textwidth]{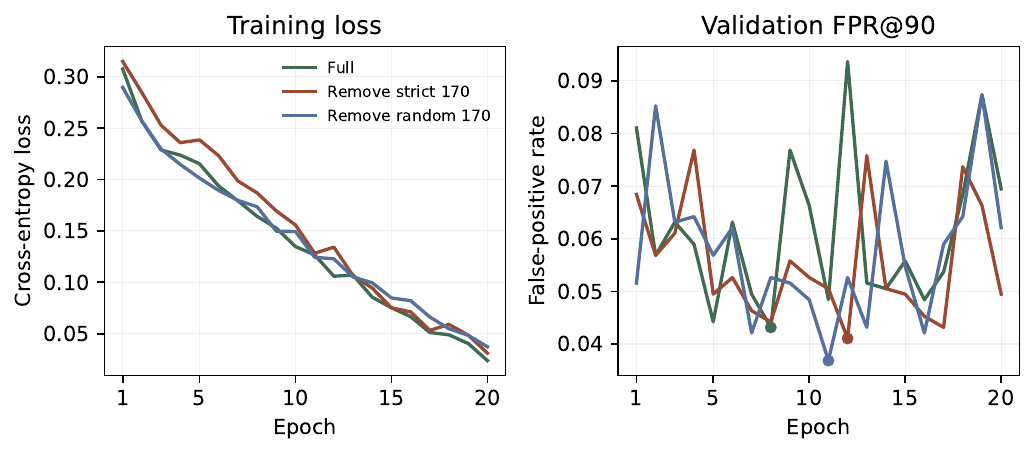}
  \caption{Training convergence for the full, strict-removal, and seed-42 matched-random-removal conditions. Dots in the validation panel mark the checkpoints selected by the prespecified validation FPR@90 rule. All runs continued through epoch 20.}
  \label{fig:app-convergence}
\end{figure*}

Training loss continued to decline after the selected
epochs (epoch-20 losses 0.0238, 0.0311, and 0.0373,
respectively), while the prespecified validation
operating point selected earlier checkpoints.
Thus, the comparison is not an early termination
artifact. Figure~\ref{fig:app-convergence} displays
the seed-42 training trajectories. Table~\ref{tab:app-multiseed}
reports five paired training seeds for all three
conditions, while Table~\ref{tab:app-random170-seeds}
reports the complementary random-removal controls.

\section{Attribution implementation and statistics}
\label{app:attribution}

Both methods target the positive-class logit for every patch, including
negatives. GradCAM++ hooks the final 3D convolution (512 channels at $4^3$
spatial resolution), applies the GradCAM++ second/third-gradient weighting,
ReLU, trilinear upsampling to $64^3$ with
\texttt{align\_corners=False}, and per-map min--max normalization; zero-span
maps become zero.

IG uses a zero input baseline and 100 trapezoidal intervals, processed in
chunks of 10. Signed IG is used for the completeness calculation; absolute IG
is min--max normalized for spatial summaries. If $f_1$ is the positive logit,
the recorded relative gap is
\[
 \frac{\left|\sum_v \mathrm{IG}_v-\{f_1(x)-f_1(0)\}\right|}
 {|f_1(x)-f_1(0)|+10^{-8}}.
\]
Across 7,742 patches the median, 95th, and 99th percentile gaps are 0.00590,
0.01439, and 0.02064. Fifteen patches exceed 0.05 and remain explicitly
recorded rather than silently excluded.

Let $d(v)$ be Euclidean distance from voxel center $(31.5,31.5,31.5)$. The
inner mask is $I=\{v:d(v)\leq r\}$ with
$r=32(0.60)^{1/3}=26.99$ voxels (82,136 voxels; 31.3\% of the cube); the outer
mask is its complement (180,008 voxels). For normalized attribution map $A$,
\[
 \mathrm{RAR}=\frac{\operatorname{mean}_{v\in O}A_v}
                    {\operatorname{mean}_{v\in I}A_v}.
\]
RAR is undefined when the inner mean is at most $10^{-12}$. RAR below one
denotes greater central attribution density. The bounded outer-density
statistic is $B=\bar A_O/(\bar A_O+\bar A_I)=\mathrm{RAR}/(1+\mathrm{RAR})$
when defined; its equal-density null is 0.5. It is not a
voxel-count-weighted fraction of total attribution.

Radial slope is an input-intensity statistic, not an attribution statistic.
The preprocessed patch is divided into eight shells with edges uniformly
spanning radii 0--32 voxels. Mean intensity is computed per shell, and ordinary
least squares fits shell mean against shell-center radius. The reported slope
has units of normalized intensity per voxel.

\begin{table*}[t]
  \caption{Extended clean-checkpoint attribution statistics over all splits.
  Entries are median [IQR]. IG quantities exclude undefined rows. $\rho$ is
  the two-sided Spearman correlation between paired GradCAM++ and IG bounded
  values.}
  \label{tab:app-attribution}
  \centering
  \scriptsize
  \setlength{\tabcolsep}{3pt}
  \begin{tabular}{lrrrrrr}
    \toprule
    Class & $n$ & RAR & Rad.\ slope & GC++ $B$ & IG $B$ & $\rho$ ($n$) \\
    \midrule
    Positive & 1,201 & 0.659 [0.492, 0.779] & $2.03{\times}10^{-5}$ [0, $8.22{\times}10^{-4}$] & 0.397 [0.330, 0.438] & 0.214 [0.185, 0.313] & $-0.283$ (969) \\
    Frangi & 3,576 & 0.761 [0.634, 0.876] & $-0.0153$ [$-0.0225$, $-0.00848$] & 0.432 [0.388, 0.467] & 0.157 [0.111, 0.209] & 0.559 (3,570) \\
    Hard & 1,787 & 0.825 [0.740, 0.925] & $-0.0220$ [$-0.0257$, $-0.0169$] & 0.452 [0.425, 0.481] & 0.195 [0.168, 0.217] & 0.485 (1,787) \\
    Easy & 1,178 & 0.713 [0.603, 0.837] & $-1.72{\times}10^{-4}$ [$-0.00353$, 0.00202] & 0.416 [0.376, 0.456] & 0.185 [0.138, 0.252] & 0.423 (1,178) \\
    \bottomrule
  \end{tabular}
\end{table*}

On the frozen training pool used for score construction, the corresponding
positive/negative associations are $\rho=-0.268$ ($n=698$,
$p=5.99\times10^{-13}$) and $\rho=0.526$ ($n=4,619$, numerical
$p<10^{-300}$). Across all splits they are $\rho=-0.283$ for positives
($n=969$) and $\rho=0.526$ for pooled negatives. These are descriptive
proxy--proxy associations, not causal attribution-sensitivity validation.

\begin{figure*}[t]
  \centering
  \includegraphics[width=0.86\textwidth]{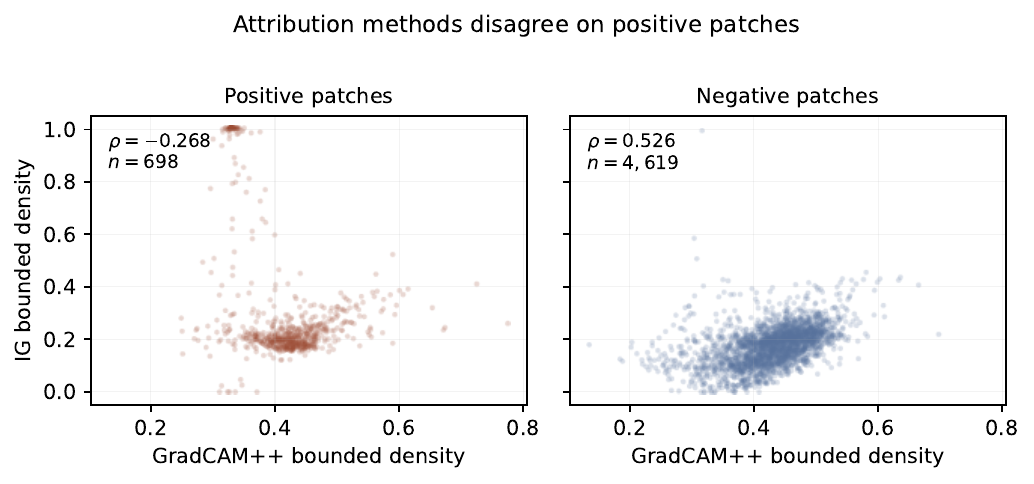}
  \caption{Training-pool agreement between GradCAM++ and IG bounded-density
  summaries. At most 2,500 points are displayed per panel; correlations use
  all finite pairs. This training-pool display concerns construction of a
  curation score and is not independent held-out evidence.}
  \label{fig:app-method-agreement}
\end{figure*}

\begin{figure*}[t]
  \centering
  \includegraphics[width=0.76\textwidth]{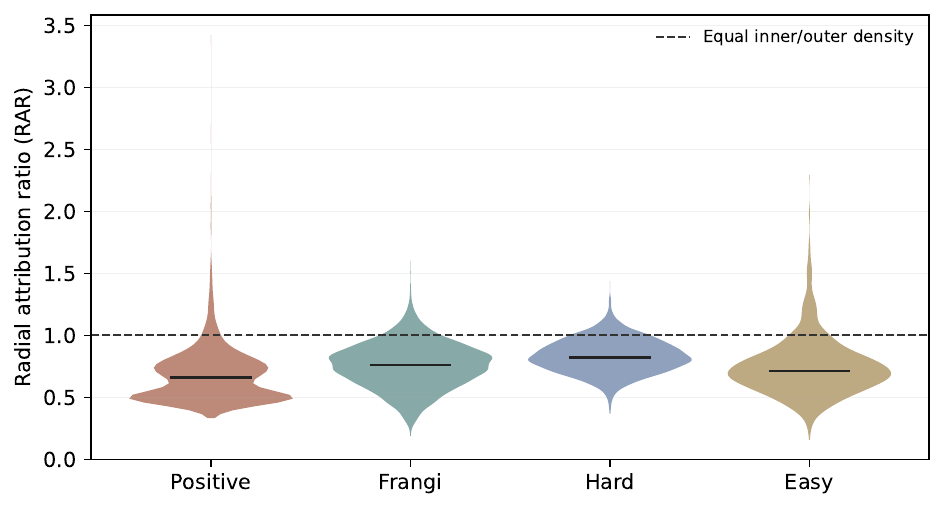}
  \caption{All-split descriptive clean-checkpoint RAR distributions by patch class. The dashed line is the equal inner/outer density reference, RAR$=1$; black bars mark class medians. Values above the visible range remain included in all numerical summaries.}
  \label{fig:app-rar-distributions}
\end{figure*}

\section{Evaluation, linkage, and statistical protocol}
\label{app:evaluation}

\begin{figure*}[t]
  \centering
  \includegraphics[width=0.88\textwidth]{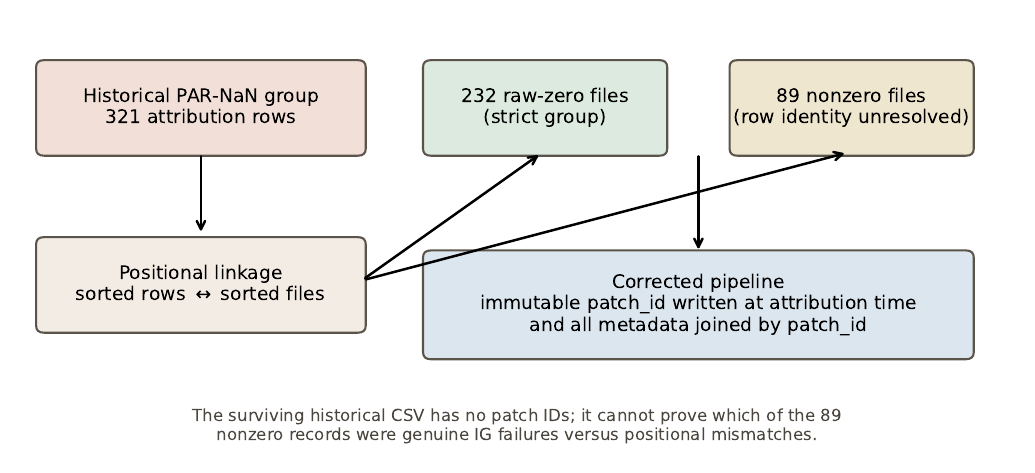}
  \caption{Historical positional linkage and the corrected identifier-preserving
  audit. Direct raw-array inspection establishes 232 zero-valued positives.
  The positional reconstruction associates the remaining 89 members of the
  broad 321-record historical group with nonzero files, but identifier-free
  historical records prevent definitive row-level reconstruction.}
  \label{fig:join-audit}
\end{figure*}

The old attribution scripts trained a mini-model on the available patch pool
and then drew a reported ``validation'' sample from the same file lists; there
was no patient-held-out split. The new protocol freezes patient-disjoint
train/validation/test membership before fitting. Model selection uses
validation only, and all conditions evaluate the same 1,136 test IDs. The
exact overlap fraction in the historical ad hoc sample is not reconstructible
because the old attribution CSV omitted identifiers, but the code establishes
that its evaluation source pool was also used for training.

The historical positive attribution CSV has columns
\texttt{class}, \texttt{par}, \texttt{radial\_slope}, and
\texttt{completeness\_gap}, but no patch ID. A downstream script reconstructed IDs
by
\[
\texttt{zip(positive\_CSV\_rows,\allowbreak \newline \ sorted(positive\_NPZ\_files))}
\]
The corrected pipeline writes the canonical relative NPZ path as
\texttt{patch\_id} into every row and performs keyed joins:
\[
\texttt{size\_by\_id[Path(attribution.\allowbreak patch\_id).stem]}
\]
For the current files, all 1,201 positive IDs match the size table uniquely
with no missing or extra IDs. Direct QC1 audit finds 0/880 finite-PAR
historical Mode-1 files raw-zero and 232/321 PAR-NaN Mode-2 files raw-zero;
the remaining 89 are nonzero. No surviving historical script computes the
claimed identical HU distributions. Moreover, the surviving source CSV
contains 232 exact-zero positive slopes while its downstream JSON serializes
324, so the historical per-row slope provenance is internally inconsistent.

For small-size enrichment, the primary $2\times2$ Fisher test is two-sided and
patch-level. The odds-ratio CI is the log-Woolf interval. Because 1,201 positive
patches arise from 957 patients, a seed-42 patient-cluster bootstrap (20,000
resamples) was also computed: median OR 2.29, 95\% percentile interval
1.44--4.13; only 0.015\% of bootstrap ORs were at or below one. This sensitivity
analysis preserves the association but does not turn patch estimates into
independent clinical observations. Spearman tests are two-sided. The primary test-split GradCAM++ RAR
intervals use 10,000 patient-cluster bootstrap resamples with seed 42;
all-split attribution summaries in Table~\ref{tab:app-attribution} are
patch-level descriptive medians and IQRs. No multiple-comparison correction is
applied to the prespecified enrichment test; method-agreement and classwise
summaries are labeled descriptive.

\begin{table}[t]
  \caption{Corrected strict Mode-2 small-aneurysm contingency tables.
  Percentages are within rows.}
  \label{tab:app-enrichment}
  \centering
  \small
  \begin{tabular}{llrr}
    \toprule
    Scope & Group & $<3$\,mm & $\geq3$\,mm \\
    \midrule
    All & Strict raw-zero & 204 (87.9\%) & 28 (12.1\%) \\
        & Non-strict & 738 (76.2\%) & 231 (23.8\%) \\
    \addlinespace
    Train & Strict raw-zero & 146 (85.9\%) & 24 (14.1\%) \\
          & Non-strict & 530 (75.9\%) & 168 (24.1\%) \\
    \bottomrule
  \end{tabular}
\end{table}

For all positives, OR=2.28 (95\% CI 1.50--3.48; Fisher
$p=5.71\times10^{-5}$). In train, OR=1.93 (1.21--3.07; $p=0.00522$). These
replace the old OR approximately 2.88. The size covariate is an existing
estimate: spacing comes from NIfTI for 1,200/1,201 rows; 1,007 single-slice
cases use a one-slice $z$-height fallback and ``Z only'' dome estimate, while
194 use DICOM image-position-derived height and $\max(XY,Z)$. It is not a new
blinded clinical remeasurement.

\section{Curation controls and operating points}
\label{app:curation}

The full-positive baseline retains 868 positives, 868 Frangi negatives, 868 hard
negatives, and 856 easy negatives. Strict removal deletes all 170 train
positives satisfying the frozen strict flag. Five matched controls use seeds
42--46 to independently permute the 868 positives and remove the first 170.
Every removal arm retains 698 positives and every negative. Validation and
test IDs are identical across conditions. The seed controls both sample
selection and training, so their dispersion describes the combined variation
of the randomized control pipeline rather than selection variance alone.

\begin{table*}[t]
  \caption{Fixed-test operating points for the three positive-removal
  conditions. Each FPR is the minimum empirical FPR attaining at least the
  stated sensitivity.}
  \label{tab:app-operating}
  \centering
  \small
  \setlength{\tabcolsep}{4pt}
  \begin{tabular}{lrrrrrrr}
    \toprule
    Condition
      & AUC
      & \multicolumn{2}{c}{@80\% sens.}
      & \multicolumn{2}{c}{@90\% sens.}
      & \multicolumn{2}{c}{@95\% sens.} \\
    \cmidrule(lr){3-4}
    \cmidrule(lr){5-6}
    \cmidrule(lr){7-8}
    & & FPR & Achieved & FPR & Achieved & FPR & Achieved \\
    \midrule
    Full-positive
      & 0.9671
      & 0.0310 & 0.8036
      & 0.0661 & 0.9048
      & 0.1064 & 0.9524 \\
    Strict 170
      & 0.9589
      & 0.0341 & 0.8095
      & 0.0671 & 0.9048
      & 0.2366 & 0.9524 \\
    Random 170 (s42)
      & 0.9508
      & 0.0351 & 0.8036
      & 0.1012 & 0.9048
      & 0.2335 & 0.9524 \\
    \bottomrule
  \end{tabular}
\end{table*}


\begin{table}[t]
  \caption{Five exact matched random-positive controls.
  Each run removes 170 of 868 training positives
  and no negatives; validation and test membership
  are frozen.}
  \label{tab:app-random170-seeds}
  \centering
  \footnotesize
  \setlength{\tabcolsep}{3pt}
  \begin{tabular}{@{}lccc@{}}
    \toprule
    Seed & \shortstack{Selected\\epoch}
         & Test AUC & \shortstack{Test\\FPR@90} \\
    \midrule
    42 & 11 & 0.9508 & 0.1012 \\
    43 & 10 & 0.9705 & 0.0950 \\
    44 & 6  & 0.9665 & 0.0816 \\
    45 & 10 & 0.9656 & 0.0579 \\
    46 & 9  & 0.9569 & 0.0961 \\
    \midrule
    Mean $\pm$ SD & --- &
    $0.9620\pm0.0080$ & $0.0864\pm0.0175$ \\
    \bottomrule
  \end{tabular}
\end{table}

Strict removal has lower FPR@90 than the matched-random mean but slightly
lower AUC, making the specificity comparison metric-dependent. The baseline
retains the best point estimate. For strict minus full AUC, a paired
patient-cluster bootstrap over the 267 test patients (20,000 resamples, seed
42) gives $-0.00820$ with 95\% percentile interval $-0.01918$ to $0.00055$.
The interval narrowly includes zero; it neither resolves the difference from
zero nor establishes equivalence.
\paragraph{Paired-seed replication.}
To separate training-seed variation from the deletion effect, a second
experiment trains all three conditions with five paired seeds (42--46),
holding the removed-sample selection fixed. Table~\ref{tab:app-multiseed}
reports per-seed results. The original selection-variance control
(Table~\ref{tab:app-random170-seeds} above) varies the removed subset under
a single training seed; the paired-seed design complements it by varying
the training seed under a single removal subset.

\begin{table*}[t]
  \caption{Five paired training seeds for all three positive-removal
  conditions. Each seed controls weight initialization; the removed-sample
  selection is fixed (selection seed 42 for the random arm). Validation and
  test membership are frozen.}
  \label{tab:app-multiseed}
  \centering
  \small
  \begin{tabular}{llrr}
    \toprule
    Condition & Seed & Test AUC & Test FPR@90 \\
    \midrule
    Full data & 42 & 0.9671 & 0.0661 \\
    Full data & 43 & 0.9689 & 0.0589 \\
    Full data & 44 & 0.9661 & 0.0909 \\
    Full data & 45 & 0.9651 & 0.0723 \\
    Full data & 46 & 0.9586 & 0.0764 \\
    \addlinespace
    Full data & Mean $\pm$ SD & $0.9652\pm0.0039$ & $0.0729\pm0.0120$ \\
    \midrule
    Strict removal & 42 & 0.9589 & 0.0671 \\
    Strict removal & 43 & 0.9581 & 0.0682 \\
    Strict removal & 44 & 0.9657 & 0.0775 \\
    Strict removal & 45 & 0.9671 & 0.0878 \\
    Strict removal & 46 & 0.9585 & 0.0816 \\
    \addlinespace
    Strict removal & Mean $\pm$ SD & $0.9617\pm0.0044$ & $0.0764\pm0.0088$ \\
    \midrule
    Random removal & 42 & 0.9508 & 0.1012 \\
    Random removal & 43 & 0.9612 & 0.1002 \\
    Random removal & 44 & 0.9568 & 0.0775 \\
    Random removal & 45 & 0.9627 & 0.0795 \\
    Random removal & 46 & 0.9510 & 0.0806 \\
    \addlinespace
    Random removal & Mean $\pm$ SD & $0.9565\pm0.0056$ & $0.0878\pm0.0118$ \\
    \bottomrule
  \end{tabular}
\end{table*}

The paired strict-minus-full AUC difference across five seeds is $-0.0035$
(paired-seed bootstrap 95\% CI $-0.0082$ to $0.0007$). Strict removal has
higher AUC in one of five paired seeds (seed 45). The confidence interval
includes zero; targeted deletion shows no reproducible advantage.

\begin{table}[t]
  \caption{Clean-test evaluation. The seed-42 models are re-evaluated on the
  test set after removing the 34 raw-zero positive test patches ($n=1102$).
  No retraining is performed.}
  \label{tab:app-cleantest}
  \centering
  \small
  \begin{tabular}{lrr}
    \toprule
    Condition & Clean AUC & Clean FPR@90 \\
    \midrule
    Full data & 0.9589 & 0.0692 \\
    Strict removal & 0.9486 & 0.1167 \\
    \addlinespace
    Strict $-$ Full & $-0.0103$ & --- \\
    95\% CI & $-0.0239$ to $0.0006$ & --- \\
    \bottomrule
  \end{tabular}
\end{table}

The strict model's FPR@90 nearly doubles on the clean test set (0.0671 to
0.1167), while the full model's remains stable (0.0661 to 0.0692). The
strict-minus-full AUC gap widens from $-0.0035$ (multi-seed mean) to
$-0.0103$ on the clean test. This result contradicts the hypothesis that
full-data performance is inflated by learning a ``blank equals positive''
shortcut.

\section{Reproducibility}
\label{app:reproducibility}

\begin{table*}[t]
  \caption{SHA256 manifest. Hashes identify exact split membership,
  attribution inputs/outputs, and condition definitions.}
  \label{tab:app-hashes}
  \centering
  \scriptsize
  \begin{tabular}{p{0.28\linewidth}p{0.65\linewidth}}
    \toprule
    Artifact & SHA256 \\
    \midrule
    Manifest mapping & \nolinkurl{dd11fed82c60a86a3bbb3704359147803c0267d498dde952fee64b4ad1a5456c} \\
    Train ID list & \nolinkurl{26a0bd83e7e9bffd0e859fcc5331f10a4b7f4f951b59564a82f91536d561d1b6} \\
    Validation ID list & \nolinkurl{553c6d4ca73996af211b9d0d8b8455e24c4f076ec805bbaade1dd71982568784} \\
    Test ID list & \nolinkurl{f9e8f98a2bcbe4eecdd8d22e0154cbb6ece531fd8b3ee1691d15578859d443c8} \\
    Manifest CSV file & \nolinkurl{2a28f92957fd644c1f2e0330bc17e3f7e9f3f18dfc31d0f7796848d9bac996de} \\
    Attribution checkpoint & \nolinkurl{753b880a408b5d7f46fb6e1cc36f376c20e0cb9fe3291c08f7363c0a6f0eadf5} \\
    Joined attribution CSV & \nolinkurl{4369142346d21e02d0cda4e7c115322aed884c09c95ffbd42f01028a4d8ca9c2} \\
    GradCAM++ CSV & \nolinkurl{476edddaa48db61401f11a29e691c402e2cc1aba72e9d9ff1f78e90bf343482d} \\
    IG CSV & \nolinkurl{357003cefedca5841c2d999733c5e7fd9b7b1750c16683438e28e60c39392579} \\
    Initial grid configuration & \nolinkurl{0b0f3b85aa443c9126651e77b1f88b76ea58428bf12742fed297e30f0c1388e6} \\
    Matched-positive configuration & \nolinkurl{0bd31394b93950da780521c172cd1fe515201aa985ad27e19256e47715e22ab3} \\
    Multi-seed matched configuration & \nolinkurl{571e666869b7febad8de9d63d313c4d7528523b98ea954bf6f187312932f92e2} \\
    Multi-seed summary & \nolinkurl{a03fb0555c6bc2615cda39e8f55fe8b6b0b8b964b83c59b5d732c2202c536778} \\
    AUC-difference bootstrap & \nolinkurl{3ea3b5b8e48bbce0f3139fabae03ad9d07cc126c14cb6de8c7bec895d171d3a8} \\
    \bottomrule
  \end{tabular}
\end{table*}

The environment is Python 3.10.20, PyTorch 2.7.1+cu128, CUDA 12.8, MONAI
1.3.2, NumPy 1.26.4, and SciPy 1.15.3. GradCAM++ and IG are custom PyTorch
implementations; Captum is not used. Deterministic seeds are set for Python,
NumPy, and PyTorch, with cuDNN deterministic mode enabled and benchmarking
disabled.

The executable mapping is: script 11 freezes the split; 13 trains the
attribution baseline; 14 generates GradCAM++/IG; 15 enforces Phase-0 coverage,
faithfulness, raw-zero, and hash gates; 16 evaluates the fixed test protocol;
17 trains each curation condition; 18 aggregates the grid; 23 reconciles QC1;
and 24 performs the explicit-ID size-enrichment analysis. CAS/recentering
scripts report unavailable rather than fabricating a source-volume
intervention.

\end{document}